\documentclass{article}
\usepackage{ijcai21}

\usepackage{times}
\usepackage{soul}
\usepackage{url}
\usepackage[hidelinks]{hyperref}
\usepackage[utf8]{inputenc}
\usepackage[small]{caption}
\usepackage{graphicx}
\usepackage{amsmath}
\usepackage{amsthm}
\usepackage{booktabs}
\usepackage{algorithm}
\usepackage{algorithmic}
\title{Nested Named-Entity Recognition on
	Vietnamese COVID-19: Dataset and Experiments\\ 
}
\author{
Ngoc C.L\^{e}$^1$\thanks{Equal contribution.}
\and
Hai-Chung Nguyen Phung$^{1,2}$\footnotemark[1]
\and
Thu-Huong Pham Thi$^1$\and
Hue Vu$^1$\and
Phuong-Thao Nguyen Thi$^1$\and
Thu-Thuy Tran$^1$\and
Hong-Nhung Le Thi$^1$\and
Thuy-Duong Nguyen-Thi$^3$\and
Thanh-Huy Nguyen$^4$\\
\affiliations
$^1$School of Applied Mathematics and Informatics, Hanoi University of Science and Technology\\
$^2$Financial Deep Mind\\
$^3$National Economics University
$^4$Faculty of Information Technology, Hanoi Open University
\emails
lechingoc@yahoo.com,
\{hchung1997, 
phamhuong.tb2k, 
hue.hnue,
phuongthao100920, 
thuytrany17,
hongnhungkchy0411, 
thuyduong2379
\}@gmail.com,
huynt@hou.edu.vn
}

\begin{document}

\maketitle

\begin{abstract}
The COVID-19 pandemic caused great losses worldwide, efforts are taken place to prevent but many countries have failed. In Vietnam, the traceability, localization, and quarantine of people who contact with patients contribute to effective disease prevention. However, this is done by hand, and take a lot of work. In this research, we describe a named-entity recognition (NER) study that assists in the prevention of COVID-19 pandemic in Vietnam. We also present our manually annotated COVID-19 dataset with nested named entity recognition task for Vietnamese which be defined new entity types using for our system.
\end{abstract}

\section{Introduction}
Covid-19 has been a major problem of human being for the year 2020 and may be even in next years. As in May, 2021, there were more than 150 millions infected people and more than three million deaths so far. Realizing the danger of the disease, countries have taken steps to prevent its spread. In Vietnam, with measures preventing including quick tracking and isolating patients as well as their close contacts, Vietnam can be considered as one of successful experiences with low number of cases and social activities are still in normal conditions. Traceability is done by patient interview about their contacts, as well as their movements and activities. Then reports is made and sent to the CDC as medical forms. After that, patients' information is extracted from these reports to assist in preventing the spread of the disease such as: date of symptom, recent contacts and date, isolation as well as admission dates. Based on these information, it is possible to localize epidemic and promptly isolate suspected cases, helping to prevent an outbreak of epidemic. These information also give managers whole picture about the current state. However, processing these data requires a lot of time, extracting necessary information by hand may takes hours, this slow-down follow-up plan. Therefore, we propose an information extraction system that automatically recognize relevant named entities presented in medical reports.

In recent years, with the development of NLP in Vietnam, there have been many Vietnamese NER datasets, most notable are VLSP 2016 \cite{vlsp2016} and VLSP 2018 \cite{vlsp2018} and or the most recent, PhoNER\_COVID19 \cite{phoner}. Datasets with named entity types contain key information: in VLSP 2016 and VLSP 2018 have persons (PER), organization (ORG), location (LOC), and other entities (MISC). PHONER\_COVID19 has the patient code (PATIENT\_ID), person name (PERSON\_NAME), age (AGE), gender (GENDER), job (OCCUPATION), location (LOCATION), organization (ORGANIZATION), symptoms and disease ( SYMPTOM \& DISEASE), transportation (TRANSPORTATION) and date (DATE). In particular, VLSP 2018 has nested entities (i.e., hierarchically structured entities). There are also many Vietnamese NER datasets being developed to contribute to the development of NLP in Vietnam and we would like to be part of that as well.

Our dataset contains COVID-19 related information, extracted from articles compiled from reputable online news sites in Vietnam, online data portal, medical reports and annotated similarly to VLSP 2018. We also re-checked manually, removed noise and corrupted data and synthesized almost all cases in Vietnam. Our dataset is annotated with eleven different named entity types and has nested entities, related to COVID-19 patients in Vietnam, including 10271 sentences and 11128 entities. We then empirically evaluate strong baseline models on our dataset using BiLSTM and the pre-trained language models PhoBERT \cite{nguyen2020}.

In the following sections, we will cover studies related to natural language processing with the subject of COVID-19. Next, section \ref{secDataset} describes our dataset, section \ref{secExperiments} we present our experiment and in section \ref{secConclusion} we summarize the content presented.
\section{Related work}
The White House and the world's leading group of researchers shared the COVID-19 Open Research Dataset (CORD-19) as a large dataset of academic articles related to COVID-19 \cite{wang2020}. Many published studies that used CORD-19, many competitions are held on this dataset that attracted a lot of teams to join and contributed many interesting and novel solutions. LitCovid \cite{chen2020} is a curated literature hub and updated daily based on articles about COVID-19 on PubMed. Kocaman et al \cite{kocacman2020} combine pre-trained NER models of PySpark that trained on biomedical and clinical dataset in a pipeline and use it to extract knowledge and information from CORD-19. Tan et al \cite{tan2020} used BERT and GPT-2 for text summarization task on CORD-19. There are also studies of COVID-19 data from social network such as Twitter \cite{zong2020}, \cite{banda2020} or Weibo \cite{hu2020}. Some datasets regarding COVID-19 in Vietnamese published recently as well, such as \cite{nguyen2025viqa} and \cite{Le2023}.
\begin{table}
    \centering
    \renewcommand\arraystretch{1.8}
    \begin{tabular}{l|l}
        \toprule
        \toprule
        \textbf{Entity Types} & \textbf{Description} \\
        \midrule
        \midrule
        NAME & Name of patient \\
        GENDER & Gender of patient \\
        AGE & Age of patient \\
        ADDRESS & Address of patient \\
        HOS\_DATE & Date that patient was admitted to hospital \\
        SYM\_DATE & Date that symptoms appear in patient \\
        ISO\_DATE & Date that patient is isolated \\
        POS\_DATE & Date that patient positive for COVID-19 \\
        CON\_DATE & Date that patient contact with anyone\\
        CON\_PERSON & Person who contact with patient \\
        LOCATION & Location where the patient passed\\
        \bottomrule
        \bottomrule
    \end{tabular}
    \caption{Entity types description}
    \label{tblEntityDescription}
\end{table}
\section{Dataset}
\label{secDataset}
\begin{table}
    \centering
    \renewcommand\arraystretch{1.4}
    \begin{tabular}{l|rrr|r}
        \toprule
        \toprule
        \textbf{Entity Type} & \multicolumn{1}{@{}c@{\hskip0.08in}}{\hspace{0.1cm}\textbf{Train}} & \multicolumn{1}{@{}c@{\hskip0.08in}}{\hspace{0.1cm}\textbf{Valid}} & \textbf{Test} & \textbf{All} \\
        \midrule
        \midrule
        NAME & 455 & 133 & 146 & 734 \\
        AGE & 472 & 138 & 144 & 754 \\
        GENDER & 437 & 139 & 135 & 711 \\
        ADDRESS & 454 & 137 & 137 & 728 \\
        SYM\_DATE & 358 & 49 & 62 & 469 \\
        HOS\_DATE & 709 & 139 & 147 & 995 \\
        POS\_DATE & 682 & 117 & 140 & 939 \\
        ISO\_DATE & 698 & 140 & 139 & 977 \\
        CON\_PERSON & 3773 & 957 & 637 & 5367 \\
        CON\_DATE & 1448 & 336 & 262 & 2046 \\
        LOCATION & 2044 & 1067 & 987 & 4098 \\
        \hline
        Level 1 entities & 10354 & 2937 & 2579 & 15870 \\
        Level 2 entities & 848 & 352 & 301 & 1501 \\
        Level 3 entities & 282 & 51 & 42 & 375 \\
        Level 4 entities & 46 & 12 & 14 & 72 \\
        \hline
        Entities & 11530 & 3352 & 2936 & 17818 \\
        Sentences & 6162 & 2054 & 2055 & 10271 \\
        \bottomrule
        \bottomrule
    \end{tabular}
    \caption{Statistics dataset}
    \label{tblDataset}
\end{table}
It is an honor for us to be a few of scientists trusted by the COVID-19 prevention facility to share important and exclusive medical reports for COVID-19 prevention research. These reports include patient's personal information (sensitive parts encrypted), person's events (places visited, people contacted) over a certain period of time. They come from many different medical facilities, are collected at the COVID-19 prevention facility for medical team to manually extract necessary information to prevent COVID-19. Since reports have no fixed form, rule-based cannot be used to extract necessary information from them. So, we propose a NER system to extract that information automatically. Since data is still quite small, we also collected more closely similar or related data from reputable news sites and community sources. When the amount of data has been improved, we summarized and annotated them, in order to create a NER dataset to train our system's model and serve for natural language processing, artificial intelligence studies in future. Data was cleaned up by fixing spelling mistakes, typing, removing duplicates, parts unrelated to COVID-19 and obtained 10271 raw sentences. We update daily case reports that we can collect, so number of raw sentences will be different from the time we wrote this article.
\subsection{Entity Types}
COVID-19 patient timelines are critical in preventing the spread of epidemic, helping determine how long a patient can infect others (starting from when possibility of infection or symptomatic until quarantine). Based on that, tracing team can find patients contact person during that time, test and quarantine infected ones. Realizing the effectiveness of this approach, we define eleven entity types (described in Table \ref{tblEntityDescription}) which are important information to support its implementation faster in practice (instead of doing it manually at the moment). These eleven entities were defined after careful consideration in consultation with the medical team: 
\begin{itemize}
\item First, we need to extract patient information including full name (NAME), age (AGE), gender (GENDER) and address (ADDRESS).
\item Then, we need to extract above important timelines, we have divided date entity into five different date entities, including symptom onset date (SYM\_DATE), date of patient contact other (CON\_DATE), date that patient is admitted to hospital or receives treatment for COVID-19 at another medical facility (e.g. quarantine, field hospital ...) (HOS\_DATE), date of initiation of quarantine (ISO\_DATE) and date confirmed positive for COVID-19 (POS\_DATE).
\item Finally, it is necessary to extract names of people who have been in contact with patient and locations where patient has visited during the above time, to isolate and treat people and to localize the area concerned.
\end{itemize}

The challenge with our dataset is that entities are easily confused with each other, for example patient information is confused with that of others, addresses could be confused with other locations, and dates could be confused with each other, so model may misclassify entities. Similar to VLSP 2018 NER, our data has nested entities and only raw text with XML tags in our dataset. Nested entities are also a challenge in our dataset. They are ordered by medical experts, according to their importance in disease prevention (inside entity is more important than outside entity). Ordered entities make experimentation easier and more precise. Statistically, nested entities in our dataset have at most four levels. For instance:\\
Level-1 entities:\\
$<$ENAMEX TYPE=”ISO\_DATE”$>$24/8/2020$<$/ENAMEX$>$\\
Level-2 entities:\\
$<$ENAMEX TYPE=``HOS\_DATE''$>$\\
$<$ENAMEX TYPE=``ISO\_DATE''$>$\\
24/8/2020$<$/ENAMEX$><$/ENAMEX$>$\\
Level-3 entities: \\
$<$ENAMEX TYPE=``CON\_DATE''$>$\\
$<$ENAMEX TYPE=``HOS\_DATE''$>$\\
$<$ENAMEX TYPE=``ISO\_DATE''$>$\\
24/8/2020$<$/ENAMEX$><$/ENAMEX$><$/ENAMEX$>$\\
Level-4 entities: \\
$<$ENAMEX TYPE=``POS\_DATE''$>$\\
$<$ENAMEX TYPE=``CON\_DATE''$>$\\
$<$ENAMEX TYPE=``HOS\_DATE''$>$\\
$<$ENAMEX TYPE=``ISO\_DATE''$>$\\
24/8/2020\\$<$/ENAMEX$><$/ENAMEX$><$/ENAMEX$><$/ENAMEX$>$

The importance of each entity is related to medicine field so we will not present it details.
\subsection{Annotation}
Our data annotation process is divided into three phases: 
\begin{itemize}
    \item First, data from medical facilities including over 1000 patient reports were obtained. Data includes text files, pdfs and images, which were converted to text then annotated. Based on VLSP 2018, we manually annotated data with XML tags containing entity types in purpose of research supporting as VLSP 2016 and VLSP 2018. Time to annotate all data in Phase 1, including review is two weeks. At the end of Phase 1, 5154 sentences and 6481 entities have been collected.
    \item Then, we crawled through reputable online news sites containing COVID-19 keyword. These new data contains some spelling, formatting errors and some irrelevant information. We removed duplicates and normalized Vietnamese punctuation automatically. After that, unrelated parts were manually removed, typing and spelling errors were manually corrected. We spent two weeks working on it and more than a week later reviewing. After cleaning data, 7952 sentences and 7069 entities have been collected.
    \item In the last phase, we updated more data from community sources (data processing organizations, public portals, etc), normalized and corrected them. This work was completed in two weeks including review. We reviewed all the data once again in two weeks , encrypts personal and sensitive information. After finishing, we had a total of 10271 sentences and 11128 entities.
\end{itemize}

We divided dataset into three parts: train, validation, and test. Detailed statistics of dataset in Table \ref{tblDataset}.
\section{Experiments}
\label{secExperiments}
\subsection{Data processing}
We segmented sentences and words with RDRsegmenter \cite{nguyen2018} and Trankit \cite{trankit} and choose RDRsegmenter for the performance with PhoBERT. Train, validation, and test data were converted into data files in CoNLL 2003 \cite{sang2003} format. For sentence segmentation, we found that in many cases, in order to recognize entities, it needs information from several around sentences. In other words, to recognize entity in a certain sentence, we need context of paragraph containing that sentence. For example:\\
\textit{Vietnamese: “Ngay 30.8, benh nhan di lam o cong ty den 5h30 chieu. Toi cung ngay, benh nhan duoc xac dinh co tiep xuc voi benh nhan COVID-19 nen duoc dua di cach ly.”}\\
\textit{English: “On August 30, the patient went to work at the company until 5:30 pm. On the evening of the same day, the patient was identified as having contact with a COVID-19 patient and was placed in quarantine.”}\\
ISO\_DATE in above sentence is August 30, but to determine it we need to read following sentence: “On the evening of the same day, the patient was identified as having contact with a COVID-19 patient and was placed in quarantine.” 

With the ideas of \cite{luoma2020} about cross-sentence context, each sentence in our dataset is used as the main sentence in each input example. That sentence is placed at the beginning of example and its following sentences (additional sentences) are filled in, until the max sequence length is reached. Implement this input construction, our model were trained with broader context that can recognize entities in cases that we mentioned above. We will demonstrate this in experiments.

We only consider Level-1 and Level-2 entities because we have statistically found that Level-3 and Level-4 entities are too few to do an assessment. From idea in \cite{minh2018}, to handle nested entities, we also combined entity tags of a token into an entity tag, called joint. Example: 02/08/2020 there are 2 entity tags: ISO\_DATE (level-1) and HOS\_DATE (level-2), the joint tag is HOS\_DATE+ISO\_DATE. We experiment with separate entity levels and joint to compare results between them.
\subsection{Training}
As noted above, entity recognization in our dataset is not an easy task since entities are nested and easy to confuse with each other. So, we used PhoBERT, a pre-trained language models for Vietnamese which improved the state-of-the-art in multiple Vietnamese-specific NLP tasks, and NER is one of them. PhoBERT has 2 versions: $\mathrm{PhoBERT_{base}}$ and $\mathrm{PhoBERT_{large}}$ use two nearly similar architectures (optimized based on RoBERTa \cite{liu2019}), $\mathrm{PhoBERT_{base}}$ and $\mathrm{PhoBERT_{large}}$ respectively. PhoBERT is trained on a 20GB word-level Vietnamese dataset, one of the main ideas that makes PhoBERT outperform other models (while others are trained with syllable-level datasets). We expect that PhoBERT is possible to contextualize words well so that it can reduce confusion between entities and recognize them based on different contexts.

Models we used to experiment and evaluated include: BiLSTM-CRF \cite{huang2015}, $\mathrm{PhoBERT_{base}}$, $\mathrm{PhoBERT_{large}}$, $\mathrm{PhoBERT_{base}}$-CRF, $\mathrm{PhoBERT_{large}}$-CRF. We also used $\mathrm{PhoBERT_{base}}$-CRF and $\mathrm{PhoBERT_{large}}$-CRF with cross-sentence context approach to evaluated its effectiveness. We experimented with fine-tuning approach. We used customized Adam \cite{adam} optimizer of BERT \cite{devlin2018} with weight decay of 0.01, learning rate of 5e-5 and batch size of 32, in 100 training epochs, evaluate the task performance after each epoch on the validation set.
\subsection{Evaluation}
We trained models with Level-1 and Level-2 entities separately, and then combined them to compared with models that were trained with joint-tag entities. We evaluated each model with five runs based on precision, recall and F1 score to get the most accurate results.
\subsection{Results}
In this subsection we will compare results of models to choose the best model for our system. Recognizing results with Level-1 and Level-2 entities on valid set and test set are presented in Tables \ref{tblResults-4}, \ref{tblResults5}, \ref{tblResults6}, \ref{tblResults7}, respectively. Each model was run five times, with different random seeds, averaged results presented in tables. It can be seen that approaches based on $\mathrm{PhoBERT}$ bring better results than BiLSTM. Based on error analysis, we found that $\mathrm{PhoBERT_{base}}$ and $\mathrm{PhoBERT_{large}}$ could recognize most of entities, but there were still some entities (mostly confusing entities mentioned above) that both models were not recognisable. For example, in some cases, they could not differentiate between ADDRESS and LOCATION, NAME and CON\_PERSON, or dates. That lead to $\mathrm{PhoBERT_{base}}$ and $\mathrm{PhoBERT_{large}}$ results are not different too much, even in some runs, $\mathrm{PhoBERT_{base}}$ is better, but overall, $\mathrm{PhoBERT_{large}}$ gives better results. Enhancements (CRF, cross-sentence context) have improved results, especially cross-sentence context. 

As shown in tables, $\mathrm{PhoBERT_{large}}$-CRF with cross-sentence context approach got the best results and we chose it as model for our system. Results of recognizing entities are shown in Table \ref{tblResults8}. Model recognizes patient's personal information such as: name, age, gender , address with good results. Model also works well when recognizes patients visited location, people who has been in contact with patient, dates they were in contact, and date when patient has symptoms. Recognizing ISO\_DATE, HOS\_DATE, and POS\_DATE seem quite difficult. These three days overlap in many cases because case that patient is found positive with COVID-19, then isolated and hospitalized in the same day is popular, but also improved quite well with $\mathrm{PhoBERT_{large}}$-CRF with cross-sentence context.
\section{Conclusion}
\label{secConclusion}
We presented named entity recognition study on our manually annotated Vietnam COVID-19 dataset which be defined important entity types. We also presented data processing and training on models such as BiLSTM and PhoBERT, with CRF and cross-sentence context approach. We will continue to improve our results as best we can and hope that we can deploy them into practice to help the medical team and everyone in prevention pandemic. We also expect that our dataset will drive more future research on COVID-19 prevention, named entity recognition and natural language processing.
\bibliographystyle{named}
\bibliography{ijcai21}
\begin{table*}
    \centering
    \renewcommand\arraystretch{1.8}
    \begin{tabular}{l r r r r r r}
        \toprule
        \toprule
        \textbf{Models} & &
        \multicolumn{1}{@{}c@{\hskip0.08in}}{\hspace{0.1cm}\textbf{Level-1}} & & &
        \multicolumn{1}{@{}c@{\hskip0.08in}}{\hspace{0.1cm}\textbf{Joint}} &\\
         & Prec. &
        Recall & 
        F1 & 
        Prec. &
        Recall &
        F1\\
        \midrule
        \midrule
        Bi-LSTM-CRF &66.06 &75.31 & 70.38 &65.21 &68.51 & 66.82\\[5pt]
        \hline
        $\mathrm{PhoBERT_{base}}$ &81.31 &83.54 & 82.41 &74.37 &77.95 & 76.12\\[5pt]
        \hline
        $\mathrm{PhoBERT_{large}}$ &81.52 &83.65 & 82.57 &74.35 &78.25 & 76.25\\[5pt]
        \hline
        $\mathrm{PhoBERT_{base}}$-CRF &82.42 &83.12 & 82.77 &74.93 &79.32 & 77.06\\[5pt]
        \hline
        $\mathrm{PhoBERT_{large}}$-CRF &82.02 &83.78 & 82.89 &75.17 &79.79 & 77.41\\[5pt]
        \hline
        $\mathrm{PhoBERT_{base}}$-CRF &82.83 &84.51 & 83.66 &76.60 &80.71 & 78.60 \\ 
        {+cross-sentence context} &                    &                    \\
        \hline 
        $\mathrm{PhoBERT_{large}}$-CRF&82.25 &85.78 & \textbf{83.98} &76.29 &81.12 & \textbf{78.63} \\ 
        {+cross-sentence context} &                    &                    \\
        \bottomrule
        \bottomrule
    \end{tabular}
    \caption{Results (measured with Precision, Recall and F1-score) on valid set, recognizing Level-1 entities with models}
    \label{tblResults-4}
\end{table*}
\begin{table*}
    \centering
    \renewcommand\arraystretch{1.8}
    \begin{tabular}{l r r r r r r}
        \toprule
        \toprule
        \textbf{Models} & &
        \multicolumn{1}{@{}c@{\hskip0.08in}}{\hspace{0.1cm}\textbf{Level-2}} & & &
        \multicolumn{1}{@{}c@{\hskip0.08in}}{\hspace{0.1cm}\textbf{Joint}} &\\
         & Prec. &
        Recall & 
        F1 & 
        Prec. &
        Recall &
        F1\\
        \midrule
        \midrule
        {Bi-LSTM-CRF} &60.64 &64.12 & 62.33 &61.59 &66.13 & 63.78\\[5pt]
        \hline
        $\mathrm{PhoBERT_{base}}$ &68.56 &72.15 & 70.31 &67.44 &70.55 & 68.96\\[5pt]
        \hline
        $\mathrm{PhoBERT_{large}}$ &69.35 &72.65 & 70.96 &68.58 &72.62 & 70.54\\[5pt]
        \hline
        $\mathrm{PhoBERT_{base}}$-CRF &68.88 &72.79 & 70.78 &68.24 &72.38 & 70.25\\[5pt]
        \hline
        $\mathrm{PhoBERT_{large}}$-CRF &69.73 &73.22 & 71.43 &68.09 &72.66 & 70.30\\[5pt]
        \hline
        $\mathrm{PhoBERT_{base}}$-CRF &72.62 &74.14 & 73.37 &69.23 &73.39 & 71.25 \\ 
        {+cross-sentence context}                                       \\
        \hline 
        $\mathrm{PhoBERT_{large}}$-CRF &73.04 &74.23 & \textbf{73.63} &68.90 &73.85 & \textbf{71.29} \\ 
        {+cross-sentence context}            \\
        \bottomrule
        \bottomrule
    \end{tabular}
    \caption{Results (measured with Precision, Recall and F1-score) on valid set, recognizing Level-2 entities with models}
    \label{tblResults5}
\end{table*}
\begin{table*}
    \centering
    \renewcommand\arraystretch{1.8}
    \begin{tabular}{l r r r r r r}
        \toprule
        \toprule
        \textbf{Models} & &
        \multicolumn{1}{@{}c@{\hskip0.08in}}{\hspace{0.1cm}\textbf{Level-1}} & & &
        \multicolumn{1}{@{}c@{\hskip0.08in}}{\hspace{0.1cm}\textbf{Joint}} &\\
         & Prec. &
        Recall & 
        F1 & 
        Prec. &
        Recall &
        F1\\
        \midrule
        \midrule
        {Bi-LSTM-CRF} &69.46 &71.12 & 70.28 &68.60 &71.31 & 69.93\\[5pt]
        \hline
        $\mathrm{PhoBERT_{base}}$ &78.95 &81.59 & 80.25 &72.65 &74.12 & 73.38\\[5pt]
        \hline
        $\mathrm{PhoBERT_{large}}$ &78.69 &82.17 & 80.39 &72.25 &74.15 & 73.19\\[5pt]
        \hline
        $\mathrm{PhoBERT_{base}}$-CRF &80.02 &81.88 & 80.94 &72.91 &75.31 & 74.09\\[5pt]
        \hline
        $\mathrm{PhoBERT_{large}}$-CRF &80.09 &82.65 & 81.35 &73.64 &75.69 & 74.65\\[5pt]
        \hline
        $\mathrm{PhoBERT_{base}}$-CRF &80.68 &83.14 & 81.89 &74.99 &77.22 & 76.09 \\ 
        {+cross-sentence context} \\
        \hline 
        $\mathrm{PhoBERT_{large}}$-CRF &81.08 &83.63 & \textbf{82.34} &74.31 &78.13 & \textbf{76.17}\\ 
        {+cross-sentence context} \\
        \bottomrule
        \bottomrule
    \end{tabular}
    \caption{Results (measured with Precision, Recall and F1-score) on test set, recognizing Level-1 entities with models}
    \label{tblResults6}
\end{table*}
\begin{table*}
    \centering
    \renewcommand\arraystretch{1.8}   
    \begin{tabular}{l r r r r r r}
        \toprule
        \toprule
        \textbf{Models} & &
        \multicolumn{1}{@{}c@{\hskip0.08in}}{\hspace{0.1cm}\textbf{Level-2}} & & &
        \multicolumn{1}{@{}c@{\hskip0.08in}}{\hspace{0.1cm}\textbf{Joint}} &\\
         & Prec. &
        Recall & 
        F1 & 
        Prec. &
        Recall &
        F1\\
        \midrule
        \midrule
        {Bi-LSTM-CRF} &54.16 &54.44 & 54.30 &51.49 &54.12 & 52.77\\[5pt]
        \hline
        $\mathrm{PhoBERT_{base}}$ &63.06 &65.51 & 64.26 &57.81 &62.12 & 59.89\\[5pt]
        \hline
        $\mathrm{PhoBERT_{large}}$ &63.46 &65.74 & 64.58 &57.80 &61.79 & 59.73\\[5pt]
        \hline
        $\mathrm{PhoBERT_{base}}$-CRF &64.99 &65.03 & 65.01 &58.02 &62.36 & 60.11\\[5pt]
        \hline
        $\mathrm{PhoBERT_{large}}$-CRF &63.41 &67.24 & 65.27 &56.82 &62.78 & 59.65\\[5pt]
        \hline
        $\mathrm{PhoBERT_{base}}$-CRF &64.87 &69.23 & 66.98 &58.31 &63.53 & \textbf{60.81} \\ 
        {+cross-sentence context} &                    &                    \\
        \hline 
        $\mathrm{PhoBERT_{large}}$-CRF &65.09 &69.51 & \textbf{67.23} &58.03 &63.23 & 60.52 \\ 
        {+cross-sentence context} &                    &                    \\
        \bottomrule
        \bottomrule
    \end{tabular}
    \caption{Results (measured with Precision, Recall and F1-score) on test set, recognizing Level-2 entities with models} 
    \label{tblResults7}
\end{table*}
\begin{table*}
    \centering
    \renewcommand\arraystretch{2}
    \begin{tabular}{l|l|r r r r r r r r r r r}
        \toprule
        \toprule
        \textbf{Dataset} &
        \textbf{Measure} &
        \multicolumn{1}{@{}c@{\hskip0.08in}}{\hspace{0.1cm}\textbf{NAM.}} & 
        \multicolumn{1}{@{}c@{\hskip0.08in}}{\hspace{0.1cm}\textbf{AGE}} &
        \multicolumn{1}{@{}c@{\hskip0.08in}}{\hspace{0.1cm}\textbf{GEN.}} &
        \multicolumn{1}{@{}c@{\hskip0.08in}}{\hspace{0.1cm}\textbf{ADD.}} &
        \multicolumn{1}{@{}c@{\hskip0.08in}}{\hspace{0.1cm}\textbf{SYM.}} &
        \multicolumn{1}{@{}c@{\hskip0.08in}}{\hspace{0.1cm}\textbf{POS.}} &
        \multicolumn{1}{@{}c@{\hskip0.08in}}{\hspace{0.1cm}\textbf{HOS.}} &
        \multicolumn{1}{@{}c@{\hskip0.08in}}{\hspace{0.1cm}\textbf{ISO.}} &
        \multicolumn{1}{@{}c@{\hskip0.08in}}{\hspace{0.1cm}\textbf{LOC.}} &
        \multicolumn{1}{@{}c@{\hskip0.08in}}{\hspace{0.1cm}\textbf{C.DATE}} &
        \multicolumn{1}{@{}c@{\hskip0.08in}}{\hspace{0.1cm}\textbf{C.PER.}} \\
        \midrule
        \midrule
         & \textbf{Prec.} & 99.08 & 97.69 & 98.31 & 99.87 & 83.29 & 71.17 & 76.79 & 74.78 & 77.97  & 86.02 & 85.67 \\
        \textbf{Valid} & \textbf{Recall} & 99.93 & 99.38 & 99.99 & 98.46 & 87.25 & 74.21 & 78.59 & 78.18 & 82.39  & 87.59 & 86.19 \\
         & \textbf{F1} & 99.50 & 98.53 & 99.15 & 99.16 & 85.23 & 72.66 & 77.68 & 76.44 & 80.12  & 86.80 & 85.93 \\
        \hline
         & \textbf{Prec.} &95.18 &98.47 &99.89 &98.62 &74.29 &74.54 &75.85 &78.54 &79.20 &85.45 &84.83 \\
        \textbf{Test} & \textbf{Recall} &96.12 &99.31 &99.65 &98.72 &75.42 &77.89 &79.12 &81.20 &83.60 &87.19 &88.24 \\
         & \textbf{F1} & 95.65 & 98.89 & 99.77 & 98.67 & 74.85 & 76.18 & 77.45 & 79.85 & 81.34  & 86.31 & 86.50\\
        \bottomrule
        \bottomrule
    \end{tabular}
    \caption{Results (measured with Precision, Recall and F1-score) on valid, test set entities, recognizing with $\mathrm{PhoBERT_{large}}$-CRF with cross-sentence context approach}
    \label{tblResults8}
\end{table*}

\bibliographystyle{plain}

\end{document}